\pdfoutput=1

\documentclass[11pt]{article}
\usepackage{authblk}

\usepackage{acl}
\usepackage{times,latexsym}
\usepackage{url}
\usepackage[T1]{fontenc}

\usepackage{graphicx}
\usepackage{amsmath}
\usepackage{tabularx}
\usepackage{multicol}
\usepackage{multirow}
\usepackage{algorithm}
\usepackage{algpseudocode}
\usepackage{inconsolata}
\usepackage{tikz}
\usepackage{tcolorbox}
\usepackage{enumitem}
\usepackage{booktabs}
\usepackage{lineno}

\usepackage{xspace,mfirstuc,tabulary}

\usepackage[utf8]{inputenc}

\usepackage{microtype}

\newcounter{notecounter}
\newcommand{\enotesoff}{\long\gdef\enote##1##2{}}
\newcommand{\enoteson}{\long\gdef\enote##1##2{{
			\stepcounter{notecounter}
			{\large\textbf{ \hspace{1cm}\arabic{notecounter} $<<<$ ##1: ##2 $>>>$\hspace{1cm}}}}}}
\enoteson
\enotesoff

\newcommand{\tinypm}{\scriptscriptstyle\pm}

\newtcbox{\inlinepattern}{on line,colback=c0_new!10,colframe=white,size=fbox,arc=3pt, box align=base,before upper=\strut,
	top=-4pt, bottom=-4pt, boxrule=0pt}
\newtcbox{\pattern}{on line,colback=c0_new!10,colframe=white,size=fbox,arc=3pt, box align=base,before upper=\strut,
	top=-2pt, bottom=-2pt, boxrule=0pt}
\newtcolorbox{multipattern}{on line,colback=c0_new!10,colframe=white,size=fbox,arc=3pt, box align=base, top=-2pt, bottom=0pt, boxrule=0pt, before=\adjustbox{valign=c}\bgroup, after=\egroup, before upper=\strut}

\definecolor{c0}{cmyk}{1,0.3968,0,0.2588}
\definecolor{c0_new}{cmyk}{0.48,0.0,0.28,0.75}
\definecolor{c1}{cmyk}{0,0.6175,0.8848,0.1490} 
\definecolor{c2}{cmyk}{0.1127,0.6690,0,0.4431} 
\definecolor{c3}{cmyk}{0.6765,0.2017,0,0.0667} 
\definecolor{c4}{cmyk}{0.3081,0,0.7209,0.3255} 
\definecolor{c5}{cmyk}{0,0.8765,0.7099,0.3647} 
\definecolor{cwhite}{cmyk}{0,0,0,0}
\definecolor{darkgrey}{RGB}{180,180,180}
\definecolor{decentgrey}{RGB}{220,220,220}
\usetikzlibrary{calc,fit,positioning,arrows,intersections}
\newcommand\mask{\textit{MASK}}

\newcommand\ourmethod{MEAL\xspace}
\newcommand\almethod{IPUSD\xspace}
\newcommand\ppkl{Prompt-Pair-KL\xspace}
\newcommand\PPKLshort{PP-KL}
\newcommand\ppklshort{pp-kl}
\newcommand\runissue{run variability\xspace}
\newcommand\Runissue{Run variability\xspace}
\newcommand\RunIssue{Run Variability\xspace}
\newcommand\dataissue{data selection\xspace}

\newcommand\DataIssue{Data Selection\xspace}
\newcommand{\linefont}[1]{\texttt{#1}}

\title{\ourmethod: Stable and Active Learning for Few-Shot Prompting}

\author[*$\diamond$]{Abdullatif Köksal}
\author[$\dag$]{Timo Schick}
\author[*$\diamond$]{Hinrich Sch\"utze}

\affil[*]{Center for Information and Language Processing, LMU Munich}
\affil[$\diamond$]{Munich Center of Machine Learning}
\affil[$\dag$]{Meta AI Research \protect\\
	\texttt{akoksal@cis.lmu.de}}

\def\mathindent{\phantom{ab}}
\def\mathindenttiny{\phantom{a}}
\def\mathlinebreak{\\[0.1cm]}

\begin{document}
\maketitle

\begin{abstract}
	
	\enote{ts}{the story should be:  [[(1) prompt-based fsl suffers from high variance due to seeds and data selection (2) this is an issue because ... (3) to fix this problem, we propose two new methods -- one for run variability, and one for data selection (4) our approaches alleviate the issue]] I think the abstract should talk a bit more about (2) }
	Few-shot classification has made great
	strides due to foundation models
	that, through priming and prompting, are highly effective
	few-shot learners. However, this approach has high variance
	both across different sets of few shots (\emph{data selection}) and across different
	finetuning runs (\emph{run variability}). This is problematic not only because it impedes the fair comparison of different approaches, but especially because it makes few-shot learning too unreliable for many real-world applications. 
	To alleviate these issues, we make two contributions for more stable and effective
	few-shot learning: First, we propose novel ensembling
	methods and show that they substantially reduce \emph{run variability}.
	Second, we 
	introduce a new active learning (AL) criterion for \emph{data selection} and present the
	first AL-based approach specifically tailored towards prompt-based learning. In our experiments,
	we show that our combined method, \ourmethod
	(\textbf{M}ultiprompt finetuning and prediction
	\textbf{E}nsembling with \textbf{A}ctive \textbf{L}earning), 
	improves overall performance of prompt-based finetuning
	by 2.3 points on five diverse tasks. We publicly share our code and data splits
	in \url{https://github.com/akoksal/MEAL}.
\end{abstract}

\section{Introduction}
Pretrained language models (PLMs) are effective
few-shot learners when conditioned with
a few examples in the input  \cite[i.a.]{gpt3, sewon-etal-2022-rethinking}
or finetuned with a masked language modeling objective on samples converted into cloze-style phrases
\cite{schick-schutze-2021-exploiting, gao-etal-2021-making}. 
Prompt-based finetuning is especially promising  as it
enables researchers to train relatively small models as few-shot classifiers that can make accurate predictions
with a minimal investment of time and effort.

\begin{figure}
	\centering
	\includegraphics[width=.95\linewidth]{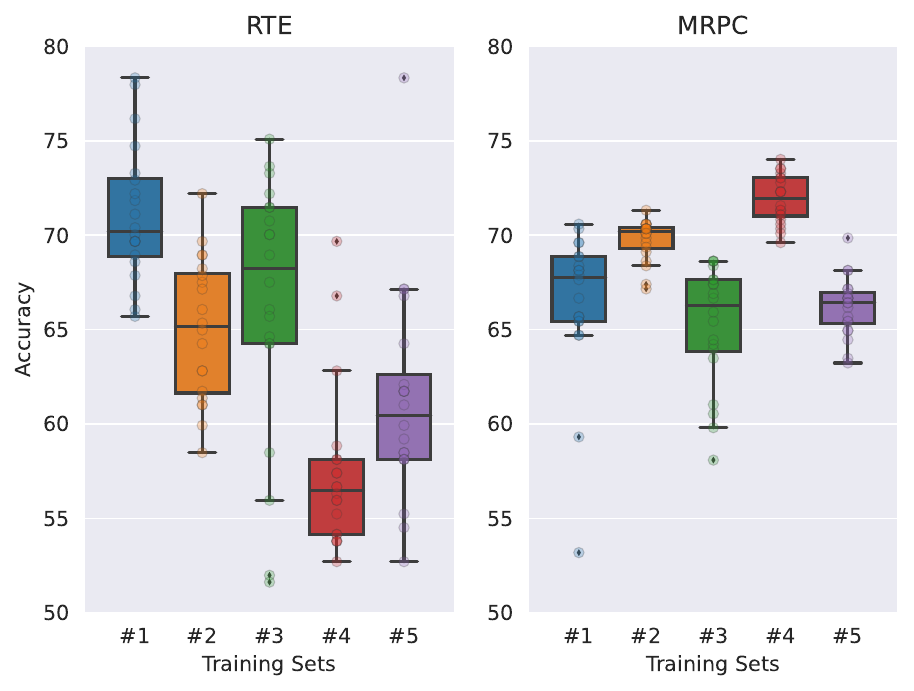}
	\caption{Multiprompt results with 32 examples for ALBERT 
		on
		RTE and MRPC. Prompt-based finetuning has \emph{large variance depending on training \dataissue
		and random initialization}.
		 The accuracy difference can be up to 23.5 with different random seeds (RTE \#3) and 13.7 with different training sets (RTE \#1 vs \#4).}
	\label{fig:instability}
\end{figure}

However, prompt-based finetuning suffers from
high variance. We observe two causes in our experiments:
\textit{\runissue} (different seeds) and \textit{\dataissue}
(different training sets).
Figure \ref{fig:instability} illustrates
this for
five equal-size training sets and 20 runs for
RTE \cite{rte} and MRPC \cite{mrpc}.
Both sources of variance are of 
particular concern in few-shot
learning. We may get lucky and select a ``good''
training set. But because no dev set is available
there is also a high risk of
selecting a ``bad'' training set, resulting in much lower performance
than possible
for the available annotation budget.
In addition, \textit{\runissue} is a great methodological
problem because it means that the exact same experimental
setup (except for different random seeds,
causing
variance in the order of training examples and dropout
layers) will give
different results. This makes fair comparison of different
algorithms and architectures  difficult.

\enote{hs}{as a reader i might ask: so why don't you just
run everything five times and compute average and standard
deviation? i guess we don't have space to explain this
here. hopefully, the rest of the paper will make it clear}

We propose new approaches to few-shot learning that address both sources of variance. We first focus on \emph{\runissue} and show based on loss/accuracy surface visualizations \cite{NEURIPS2018_a41b3bb3} that \runissue in few-shot
learning is different from fully-supervised settings:
solutions proposed for finetuning PLMs \cite{DBLP:conf/iclr/MosbachAK21}
do not work for few-shot
prompt-based finetuning. Thus, we propose ensemble
techniques to stabilize finetuning for different runs. 
After mitigating the effects of  \runissue via
a more stable finetuning mechanism, 
we are able to address training \emph{\dataissue}. We modify existing active
learning (AL) algorithms and propose a novel approach for selecting training examples
that outperforms prior algorithms -- not just in terms of final accuracy, but also regarding the diversity and representativeness of selected examples. In general, we are,
to the best of our knowledge, the first to develop
AL algorithms  tailored to
prompt-based finetuning.

We combine our contributions
-- decrease  run variance and
better training sets for improved
performance and stability of few-shot classification --
in  \textbf{\ourmethod} (\textbf{M}ultiprompt finetuning and prediction 
\textbf{E}nsembling with \textbf{A}ctive \textbf{L}earning).
\ourmethod improves performance
of prompt-based finetuning by 2.3 points on five
tasks. \textbf{Contributions:}

\begin{enumerate}
\item We propose a training procedure that produces a single
  few-shot  classification model with multiple prompts on
  top of  PET \cite{schick-schutze-2021-exploiting}.
This reduces  model space complexity and improves overall performance.

\item We show that \runissue is a big problem in few-shot  classification
and conduct an exhaustive analysis of why existing solutions do not apply to few-shot
prompt-based finetuning. We propose ensemble techniques to improve run stability.

\item We  propose a novel AL method for \dataissue that outperforms prior AL work and random selection. 
Our work is the first to demonstrate that AL is
beneficial in prompt-based learning.
\end{enumerate}

\section{Related Work}
\label{sec:related}

\textbf{Few-shot classification with language model
prompting.} GPT-3 \cite{gpt3} prepends examples
as conditioning
to the input during
inference, without  parameter
updates. PET \cite{schick-schutze-2021-exploiting,
schick-schutze-2021-just} follows a similar approach
with finetuning and achieves comparable results, with fewer parameters. 
LM-BFF \cite{gao-etal-2021-making} and
ADAPET \cite{tam-etal-2021-improving} extend PET.

\noindent\textbf{Instability.}
There are two sources of instability
in finetuning PLMs for few-shot classification:
\runissue and \dataissue. \Runissue comes from
finetuning PLMs with random seeds. \citet{DBLP:conf/iclr/MosbachAK21}
and \citet{finetuning_instability_dodge_2020} show that
finetuning PLMs is an unstable process
for fully supervised
training. Recently, \citet{zheng-etal-2022-fewnlu}
demonstrated that few-shot finetuning also exhibits \runissue
 but their experiments have different training
sets in a cross-validation scenario. They do not address
how much instability comes from finetuning vs.\ \dataissue.
Our findings suggest that the instability issue
exists in few-shot training \emph{for the same training set}, and
existing solutions for fully supervised settings (e.g.,
\citet{finetuning_instability_dodge_2020}, \citet{DBLP:conf/iclr/MosbachAK21})
do not stabilize finetuning for few-shot classification. We
propose the run ensemble method to improve the stability of
few-shot classification. The second type of
instability is training \dataissue;
we target this issue with AL.

\noindent\textbf{Active Learning.} As collecting labeled
data is time-consuming and costly, AL has
been a crucial part of supervised learning \cite{cohn1996active, settles2009active,
	 rotman2022multi}. Apart
from efficiency in data labeling, we show that some training
sets have significantly worse performance than others for
few-shot  classification. 
\citet{zhao-etal-2021-closer}
also show that the selection of the few shots matters a lot.
Hence, we
follow an AL setup to select \textit{informative}
and \textit{diverse} training sets for few-shot 
classification. \citet{schroder-etal-2022-revisiting}
recently showed that uncertainty AL
achieves significant improvements for fully supervised
settings in PLMs. Following this, we modify a variety
of AL algorithms including prior works
such as CAL \cite{cal} and BADGE \cite{badge} for
few-shot prompt-based
finetuning, and propose a novel AL
algorithm, \almethod.

\begin{figure*}
	\centering
	\includegraphics[width=0.8\linewidth]{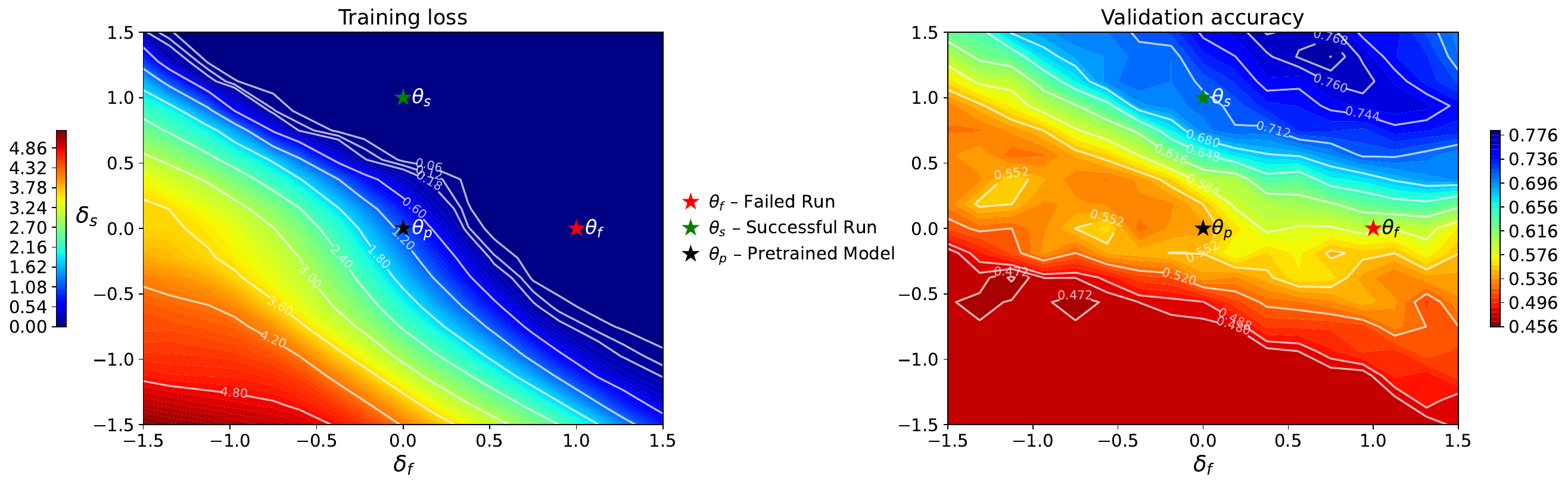}
	\caption{Loss and validation accuracy surface visualizations
		for two RTE runs with the same training set.
		Left (training loss):
		The two models
		$\theta_s$ and $\theta_f$
		have similar loss
		-- they are both located in the upper right blue zero-loss triangle.
		Right (validation accuracy):
		The successful model $\theta_s$ performs much better than the failed model
		$\theta_f$.
	}
	\label{fig:loss_surface_form}
\end{figure*}

\section{Multiprompt Finetuning}
\label{sec:multiprompt}

Let $M$ be a masked PLM, $T$ its
vocabulary, and $\textit{MASK} \in T$ the mask token. We use
Pattern-Exploiting Training (PET)
\cite{schick-schutze-2021-exploiting}
for prompt-based finetuning experiments on few-shot 
classification without knowledge distillation and unlabeled 
data. Patterns ($\mathcal{P}$) transform an input $x$ into 
a cloze-style phrase $x_p$ with a single mask. Verbalizers 
($V$) convert each label $l\in L$ into a single token 
$s_l \in T$, representing the task-specific meaning of 
the output label. 

Our prediction for a label is its probability,
according to the PLM,
as a substitution for the mask:\mathlinebreak
\mathindent$P(y|x) = \frac{\exp
s_m(V(y)|x_p)}{\sum_{y^*\in L}\exp  s_m(V(y^{*})|x_p)}$\mathlinebreak
where $s_m$ gives the raw score of $V(y)$ from a PLM $M$ for the \textit{MASK} position in the cloze-style phrase of the input.
Using the cross-entropy loss of $P$, PET trains a separate
model for each prompt (i.e., single prompt finetuning). In inference, it ensembles model predictions by logit averaging. 

We propose \textbf{multiprompt finetuning}, a modified PET that trains a
single model $M$ on all prompts for a 
task simultaneously. During inference time, we also use ensembling with
logit averaging across prompts. However, our approach generates 
a single finetuned model regardless of the number of
prompts. Compared to PET, this reduces runtime, memory, and
overall complexity.

\section{\RunIssue}
\label{sec:stability}

In few-shot classification, finetuning PLMs such as
ALBERT \cite{Lan2020ALBERT} with an MLM objective on samples
converted into cloze-style
phrases \cite{schick-schutze-2021-just} performs  comparably
to  much larger  GPT-3
\cite{gpt3}. 
Just as prompting methods are sensitive to data order
\cite{lu-etal-2022-fantastically} and label distributions
\cite{pmlr-v139-zhao21c}, finetuning PLMs also exhibits
sensitivity and instability as shown by
\citet{DBLP:journals/corr/abs-2002-06305} for a fully supervised setting.

We show that the instability of finetuning PLMs also exists
in few-shot prompt-based finetuning. Even though prompt-based
 finetuning does not introduce new parameters like classifier heads as
in fully supervised classification, there is variance from
dropout and training data order. We conduct experiments with
multiprompt finetuning with default PET
\cite{schick-schutze-2021-exploiting} settings without
knowledge distillation. Figure \ref{fig:instability}
shows that runs with different random seeds for the same
training set can vary by as much as 23.5
points.

\citet{DBLP:conf/iclr/MosbachAK21} suggest that longer
training with a low learning rate and warm-up reduces
\runissue of PLMs.  Their main motivation is to
avoid models ending up in suboptimal training loss
regions. However, this is not valid in few-shot prompt
tuning as the number of training examples is low, and
finetuning achieves almost zero training loss quickly. Our
initial experiments show that longer training does reduce
the standard deviation between different runs, but that it
also causes lower mean accuracy for most tasks,
of up to 7.3 points. 

In Figure \ref{fig:loss_surface_form},
we analyze \runissue, by creating a training
loss and validation accuracy surface visualization of two
RTE runs with \emph{the same training set} and multiprompt
finetuning. The failed model
$\theta_f$ (red)
achieves 58.5\% validation
accuracy while the successful model
$\theta_s$ (green)
achieves 71.5\%. The two
models only differ in finetuning random seed. The figure
illustrates the training loss and validation accuracy
surfaces for combinations of the model weights of the
pretrained model ($\theta_p$), the failed model
($\theta_f$), and the successful model ($\theta_s$). We
create a two-dimensional space based on $f(a, b) =
F(\theta_p + a\delta_f + b\delta_s)$, where
$\delta_f=\theta_f-\theta_p$, $\delta_s=\theta_s-\theta_p$,
and $F$ is loss (left) or accuracy (right). We use 16 values for $a$ and $b$ to plot loss and accuracy surface forms.

Figure \ref{fig:loss_surface_form} shows that there
is a large region with $\leq$1e-4 training loss (left graph,
dark blue) that includes
$\theta_f$ and $\theta_s$. However, most of this region is
suboptimal in terms of validation accuracy (right graph). This indicates that our instability problem differs
from fully supervised finetuning where large learning
rates often result in suboptimal training loss; in contrast, we
observe $\approx$0 training loss for each run, including
failed ones. Therefore, longer training with a low learning
rate and warm-up only leads to finetuned models ending up in
a similar region with lower variance, but it causes
suboptimal validation accuracy scores;
see \S\ref{sec:res_instability} for more details.

To overcome \runissue, we propose two ensemble models: We
ensemble the logits over runs in
ENSEMBLE\textsubscript{pred} and take the average of
parameters over runs in ENSEMBLE\textsubscript{para}. We
will show that, for five tasks, these (i) reduce the effect of failed runs
and \runissue and (ii) achieve higher accuracy than
accuracy averaged over runs.

The prediction of ENSEMBLE\textsubscript{pred} for $x$ is:\mathlinebreak
\mathindent$P(y|x) = \mbox{s}(\sum\limits_{r=1}^{R}[\sum\limits_{p\in\mathcal{P}}F_{r}(y|x_{p})]/(R*|\mathcal{P}|))$\mathlinebreak
where \mbox{s} is softmax, R is the number of runs, $\mathcal{P}$ is the
set of prompts, and $F_{r}$ gives,
for the finetuned model in run $r$,
the logit of each class for the input $x$ with  prompt
$p$.

Following  work on averaging deep
networks \cite{izmailov2018averaging},
we  average  each parameter of the
finetuned PLMs across runs, resulting in a single
model.
The prediction of
ENSEMBLE\textsubscript{para}
for $x$ is the prediction of this single model.

\section{\DataIssue}
\label{sec:active_learning}

Another important source of variance for few-shot
classification is training \dataissue. Figure
\ref{fig:instability} shows this effect:  validation
accuracy greatly varies, with a difference of up to 13.7.

\begin{figure}
	\centering
	\includegraphics[width=\linewidth]{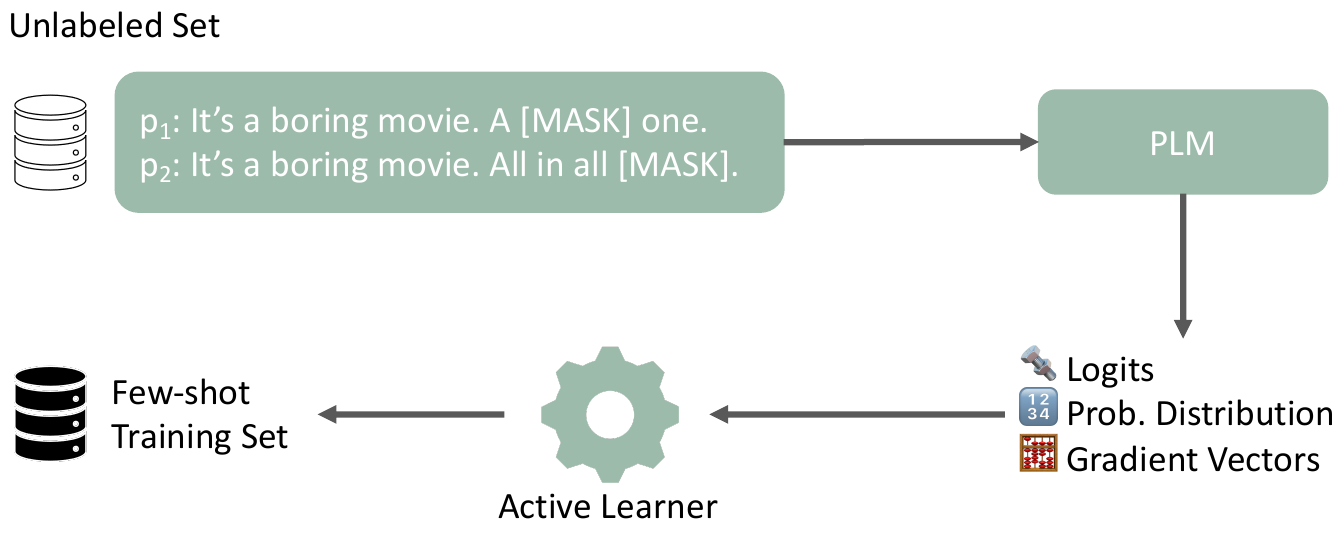}
	\caption{Our modified active learning pipeline for \dataissue 
		is illustrated with an example sentence and two prompts for sentiment analysis.
		The PLM outputs several features in a zero-shot manner. AL 
		selects a few-shot training set based on
	these output features.
	}
	\label{fig:active_learner}
\end{figure}

Figure \ref{fig:active_learner} shows how
we modify AL algorithms for \dataissue
in few-shot prompt-based finetuning.
First, we use a PLM
to get contextual embeddings, logits, and probabilities for
each unlabeled example in \emph{a zero-shot setting}.
We exploit here that, due to the cloze-style format,
PLMs can make predictions
before any finetuning.
Second, we apply modified AL
algorithms for prompts. We select all examples at once
to simplify the selection process. For each task, we select
$16*L$ training examples, where $L$ is the number of labels.

\subsection{Prior-Work Active Learning}
\label{sec:prior_al}

We use a range of prior-work AL algorithms,
including random, uncertainty-only (e.g., entropy) and
combined approaches (e.g., BADGE).
Although these are prior-work, adapting 
them to a prompt-based setup is non-trivial; e.g., for
BADGE  it requires concatenating
gradient vectors across prompts. Therefore, this 
adaptation is one of the contributions of our paper. 
Importantly, none of the prior-work leverages the prediction 
variety across different prompts. 

\noindent\textbf{Random} selection draws random examples from an unlabeled set. We report random selection results with five different seeds.

\noindent\textbf{Entropy} \cite{entropy} computes the
	entropy score of an example by summing the entropy across prompts. We then select examples with highest entropy scores.\mathlinebreak
\mathindenttiny		$\mbox{e}(x_i) \!=\! \sum \limits_{j=1}^{j=L}\sum\limits_{p\in \mathcal{P}}\!-\!P(y\!=\!l_j|x_{i,p})\ln P(y\!=\!l_j|x_{i,p})$\mathlinebreak
where $L$ is the number of labels, $\mathcal{P}$ is the set of prompts, and $x_{i,p}$ is input $x_{i}$ with pattern $p$.

\noindent\textbf{Breaking Ties (BT)} \cite{breaking_ties} selects examples with minimum difference between the highest two probability classes.\mathlinebreak
\mathindent
$\mbox{bt}(x_i)=\sum\limits_{p\in \mathcal{P}}P(y=l_{1}|x_{i,p})-P(y=l_{2}|x_{i,p})$
\mathlinebreak
where $l_{1}$ and $l_{2}$ are the labels with highest
and second highest probability for $x_{i,p}$.

\noindent\textbf{Lowest Confidence
  (LC)} \cite{lowest_confidence} calculates
\mbox{lc} as
the sum of probability scores for the predicted
class across prompts. We select examples with lowest
\mbox{lc}.
lc and bt give the same
order when there are two labels.\mathlinebreak
\mathindent $\mbox{lc}(x_i) = \sum\limits_{p\in \mathcal{P}}\max\{P(y=l_j|x_{i,p}):j=1..L\}$\mathlinebreak

\noindent\textbf{Contrastive AL (CAL)}
\cite{cal} selects examples with the highest KL divergence
between the example and its $M$ nearest
neighbors in the PLM contextual embedding space.\mathlinebreak
\mathindent
$\mbox{cal}(x_i) = \sum\limits_{m=1}^{m=M}\sum\limits_{p\in\mathcal{P}}\mbox{KL}(P(y|x_{m,p})||P(y|x_{i,p}))$\mathlinebreak

\noindent\textbf{Batch AL by Diverse Gradient
  Embeddings (BADGE)} \cite{badge}
uses as representation the
gradient of the cross entropy loss, conditioned on the one-hot
encoding of the predicted label, with respect to the parameters of
the final (output) layer. For prompt-based finetuning, we
represent $x_i$ as the concatenation of the gradient vectors across prompts
by using the decoder of the masked PLM head as
the final layer. We find
$16L$ (i.e., the number of training examples)
cluster centers
using kmeans++. These 
$16L$ cluster centers are then selected as the training set.
We average BADGE over
five seeds 
as k-means++  depends on initialization.

\subsection{Prompt-Specific Active Learning}
To make AL prior-work usable in prompt-based
learning, we sum over different prompts in \S\ref{sec:prior_al}. However,
these algorithms do not consider the varied predictions made
by the PLM
across different prompts. Therefore, we propose a new
uncertainty-only algorithm, called \textbf{\ppkl (\PPKLshort)}
 specifically designed for prompt-based
learning.
We calculate
$\mbox{\ppklshort}(x_i)$
as the sum
of KL divergence scores across prompt pairs, and
then select examples with the highest
$\mbox{\ppklshort}$.
This approach gives high scores to $x_i$ with 
high
variability in the model's predictions  indicating that such examples are ``non-redundant''
in that each prompt contributes different
information.
\mathlinebreak
\mathindent$\mbox{\ppklshort}(x_i)
= \sum\limits_{{(p, q) \in \mathcal{P}^2}}\mbox{KL}(P(y|x_{i,p})||P(y|x_{i,q}))
$\mathlinebreak

\noindent\textbf{Inter-prompt uncertainty sampling with
  diversity (\almethod)} is our novel AL
algorithm that combines prompt-specific uncertainty (i.e., \PPKLshort) 
and diversity sampling.
It first
represents each example $x$ as a vector of dimensionality
$|\mathcal{P}|\cdot|L|$, the concatenation
of the $L$ logits for $x$ for each of the patterns in
$\mathcal{P}$.
We
utilize logits here as they represent
the model's probability distribution, certainty and
divergence across different prompts.
We cluster these
representations with k-means, $k$=8.
We sample a training set,
uniformly distributed over the 8 clusters.
Then the
uncertainty score of the training set is
calculated
as the sum of its
\ppkl
scores.
We repeat
the iteration loop 1000 times. Finally,
we select the training set with the
highest uncertainty score.
We select based on 1000 iterations  to ensure a balance between randomization
and uncertainty. Our initial experiments suggest that
choosing the most uncertain
examples selects
outliers, resulting in poor performance.
As  k-means and sampling depend on 
random seed, we repeat \almethod five times.
See
\S\ref{sec:appendix_algorithm} for
the pseudo-code of \almethod.

\section{Experiments and Results}
\label{sec:experiments}

\begin{table*}
\footnotesize
	\centering
	\setlength\tabcolsep{4.75pt}
	\begin{tabular}{llccccccc}
		\toprule
		\textbf{No} & \textbf{Finetuning} & \textbf{Stability Technique} & \textbf{RTE} & \textbf{SST-2} & \textbf{SST-5} & \textbf{TREC} & \textbf{MRPC} & \textbf{Average}\\

		\midrule
		\linefont{L1} & \multirow{4}{*}{Single Prompt} & Default (PET) & 63.7$\tinypm$\tiny{1.91} &93.1$\tinypm$\tiny{0.54} &51.6$\tinypm$\tiny{0.73} &81.3$\tinypm$\tiny{1.11} &67.7$\tinypm$\tiny{1.07} &71.5$\tinypm$\tiny{1.07} \\
		\linefont{L2} & & Longer Training &  63.4$\tinypm$\tiny{0.42} &92.1$\tinypm$\tiny{0.10} &50.2$\tinypm$\tiny{0.20} &74.0$\tinypm$\tiny{0.45} &67.0$\tinypm$\tiny{0.20} &69.3$\tinypm$\tiny{0.27}\\
		\linefont{L3} & & ENSEMBLE\textsubscript{para} & 64.0$\tinypm$\tiny{1.35} &93.2$\tinypm$\tiny{0.27} &51.9$\tinypm$\tiny{0.40} &80.6$\tinypm$\tiny{0.56} &67.5$\tinypm$\tiny{0.70} & 71.4$\tinypm$\tiny{0.66}\\
		\linefont{L4} & & ENSEMBLE\textsubscript{pred} & 64.0$\tinypm$\tiny{0.95} &93.2$\tinypm$\tiny{0.30} &52.0$\tinypm$\tiny{0.51} &81.9$\tinypm$\tiny{0.72} &67.9$\tinypm$\tiny{0.58} &71.8$\tinypm$\tiny{0.61}\\
		\midrule
		\linefont{L5} & \multirow{4}{*}{Multiprompt} & Default (PET) & 64.2$\tinypm$\tiny{4.88} &91.7$\tinypm$\tiny{1.55} &52.1$\tinypm$\tiny{0.92} &82.9$\tinypm$\tiny{1.77} &67.9$\tinypm$\tiny{2.25} &71.8$\tinypm$\tiny{2.28}\\
		\linefont{L6} & & Longer Training & 66.2$\tinypm$\tiny{0.45} &93.0$\tinypm$\tiny{0.08} &50.4$\tinypm$\tiny{0.26} &79.0$\tinypm$\tiny{0.34} &66.8$\tinypm$\tiny{0.36} &71.1$\tinypm$\tiny{0.30} \\
		\linefont{L7} & & ENSEMBLE\textsubscript{para} & 64.4$\tinypm$\tiny{2.53} &92.5$\tinypm$\tiny{0.31} &52.9$\tinypm$\tiny{0.47} &83.7$\tinypm$\tiny{1.00} &68.3$\tinypm$\tiny{1.34} & 72.4$\tinypm$\tiny{1.13} \\
		\linefont{L8} & & ENSEMBLE\textsubscript{pred} & 65.4$\tinypm$\tiny{2.86} &92.4$\tinypm$\tiny{0.43} &52.8$\tinypm$\tiny{0.42} &84.4$\tinypm$\tiny{0.87} &68.9$\tinypm$\tiny{1.03} & \textbf{72.8}$\tinypm$\tiny{1.12} \\
		\bottomrule
	\end{tabular}
	\caption{Comparing stability techniques
		for prompt-based finetuning with single and
		multiple prompts with ALBERT on randomly
		selected training sets. Multiprompt
		improves overall performance
		compared to single
		prompt. ENSEMBLE\textsubscript{pred}
		improves stability while achieving higher
		performance for single prompt and multiprompt.
		Standard deviation is calculated across
		runs (trials for ENSEMBLE) and averaged over
		five random training sets. 
}

	\label{tab:stability_results}
\end{table*}

\paragraph{Setup}
\label{sec:setup}
We use a diverse set of five classification tasks to compare single to
multiprompt finetuning, analyze
\runissue, and evaluate AL
algorithms: RTE \cite{rte}, SST-2,
SST-5 \cite{sst}, TREC \cite{trec}, and MRPC \cite{mrpc}. We use
four prompts for each, described
in \S\ref{sec:appendix_datasets}. We report results on the
validation set as we conducted all experiments without 
hyperparameter tuning by assuming a realistic few-shot
scenario in which no dev set is available for tuning.\footnote{Additionally,
	RTE has no public test set. We therefore use the validation set for consistency
	across datasets.}
For single prompt and multiprompt finetuning,
we use PET's defaults
\cite{schick-schutze-2021-exploiting}. For longer training,
we compare PET's  defaults to  
\citet{DBLP:conf/iclr/MosbachAK21}'s adapted settings.
PET's defaults: 1e-5 learning
rate, 10 epochs, no warm-up.
Adapted settings:
1e-6 learning rate,  50 epochs,
linear scheduled warm-up with a 0.1 ratio.

\paragraph{Evaluation Metrics}
\label{sec:datasets}
We report the average of both accuracies and run/training
set standard deviations for a given
dataset and AL algorithm over five training
sets and five runs for each training set. For AL algorithms
without variance (e.g., entropy and \ppkl), we do not report
training set standard deviation as the algorithm outputs a
single training set. We increase the number of runs
from 5 to 20 for stability experiments and report run variance over
20 runs for default and longer training. For ENSEMBLE,
the ensemble size is 5 and
we report variance over 4 trials.

For \dataissue, we treat training examples of each dataset as unlabeled
data  from which few-shot examples are picked  via AL.
We report the
average accuracy over five datasets and the average ranking of
AL algorithms for each dataset following
\citet{schroder-etal-2022-revisiting} with additional
analysis of diversity \cite{zhdanov2019diverse}, representativeness \cite{dor2020active},
and label entropy \cite{prabhu-etal-2019-sampling} as explained in \S\ref{sec:al_additional_analysis}.

\paragraph{Single and Multiprompt Finetuning}
\label{sec:res_single_multi}
In Table \ref{tab:stability_results}, we compare single 
prompt to multiprompt finetuning. Multiprompt 
consistently outperforms single prompt on
average accuracy for each stability technique, up to 1.8
 points (\linefont{L2} vs \linefont{L6}). Across all 20 experimental
 setups for each dataset and stability technique, 
 multiprompt  (\linefont{L5} -- \linefont{L8}) achieves better average 
 accuracy in 16 cases than single prompt (\linefont{L1} --
 \linefont{L4}). 
 Even though multiprompt  produces a 
 higher standard deviation in the default setup (\linefont{L5} vs \linefont{L1}),  
 ENSEMBLE (\linefont{L7}, \linefont{L8}) overcomes this. Overall, multiprompt finetuning 
 not only provides better overall performance than single prompt, 
 but also simplifies training and deployment because it outputs a 
 single model, compared to one model per prompt for single prompt.

\paragraph{\RunIssue}
\label{sec:res_instability} For each task,
we compare the default PET \cite{schick-schutze-2021-exploiting} setup (hyperparameters given in
\S\ref{sec:setup}),
\citet{DBLP:conf/iclr/MosbachAK21}'s proposal of
longer training with a lower learning rate and warm-up
training, and ENSEMBLE over five randomly selected training sets.

Table \ref{tab:stability_results} 
shows  that \citet{DBLP:conf/iclr/MosbachAK21}'s
longer training reduces run standard
deviation, but causes suboptimal accuracy
results for SST-5, TREC, and MRPC in multiprompt
finetuning (\linefont{L6} vs \linefont{L5}), and for all datasets in single prompt
finetuning (\linefont{L2} vs \linefont{L1}). We conclude: a longer training approach is
not advisable for practical scenarios.

ENSEMBLE\textsubscript{pred} consistently reduces 
the standard deviation for each dataset, both in
single prompt (\linefont{L4} vs \linefont{L1}, 43\%) 
and multiprompt finetuning (\linefont{L8} vs \linefont{L5}, 51\%). 
This reduction in standard 
deviation is accompanied by an increase in 
accuracy of up to 1.0 absolute points,
contrary to longer training.
On the other
hand, ENSEMBLE\textsubscript{para} consistently performs better than
the default only in multiprompt (\linefont{L7} vs \linefont{L5}),
but speeds up the prediction process during
inference time with a single model while also reducing
the standard deviation.

On top of that, both
ENSEMBLE techniques avoid failed runs. For example, the
default approach with multiprompt gets 87.4\% average accuracy (not shown in 
the table) with
one of the five random training sets in SST-2 while its
worst run with the same training set
has
77.6\% accuracy. ENSEMBLE\textsubscript{pred} and ENSEMBLE\textsubscript{para} ensure 
better average accuracy (88.5\% and 88.8\%) without any suboptimal
models (accuracy of worst trials: 87.8\% and 88.4\%) for the same training set. 
Thus,
the default approach
can result in suboptimal performance and is therefore
not reliable for real-world applications without 
validation data.
Overall, ENSEMBLE\textsubscript{pred} achieves clearly better overall performance
and a lower standard deviation, but with the additional cost of multiple models
(i.e. five in our experiments)
during inference time.
ENSEMBLE\textsubscript{para} is an alternative
approach to increase stability and performance while
providing a single model with lower time complexity during inference
time.

\paragraph{\DataIssue}

\begin{table*}
	\footnotesize
	\centering
	\begin{tabular}{llcccccc}
		\toprule
		 \textbf{No} & \textbf{Algorithms} & \textbf{RTE} & \textbf{SST-2} & \textbf{SST-5} & \textbf{TREC} & \textbf{MRPC} & \textbf{Average}\\
		\midrule
		\linefont{L1} & Random & 65.3$\tinypm$\tiny{5.8} & 92.1$\tinypm$\tiny{2.4} & 52.8$\tinypm$\textbf{\tiny{0.7}} & 83.8$\tinypm$\tiny{2.3} & 69.3$\tinypm$\tiny{2.7} & 72.6$\tinypm$\tiny{2.8}\\
		\midrule
		 \linefont{L2} & Entropy & \underline{71.1} & 89.3 & 49.0 & 76.2 & 68.9 & 70.9 \\
		\linefont{L3} & Lowest Confidence & \textbf{\underline{71.8}} & 91.4 & 48.8 & 72.6 & \underline{70.1} & 70.9  \\
		\linefont{L4} & Breaking Ties & \textbf{\underline{71.8}} & 91.4 & 49.9 & 77.2 & \underline{70.1} & 72.1 \\
		 \linefont{L5} & \ppkl \textit{(Ours)} & 59.6 & 89.8 & \textbf{\underline{53.5}} & 77.4 & 65.4 & 69.1  \\
		\linefont{L6} & CAL \cite{cal} & 56.7 & \underline{92.9} & 49.0 & 81.6 & \textbf{\underline{71.8}} & 70.4\\
		\linefont{L7} & BADGE \cite{badge} & \underline{68.7}$\tinypm$\tiny{8.9} & \textbf{\underline{93.2}}$\tinypm$\textbf{\underline{\tiny{0.8}}} & 51.2$\tinypm$\tiny{2.8} & 82.7$\tinypm$\tiny{3.1} & \underline{70.3}$\tinypm$\textbf{\underline{\tiny{0.9}}} & \underline{73.2}$\tinypm$\tiny{3.3} \\
		
		\linefont{L8} & \almethod \textit{(Ours)} & \underline{70.1}$\tinypm$\textbf{\underline{\tiny{3.8}}} & \underline{92.9}$\tinypm$\textbf{\underline{\tiny{0.9}}} & 51.4$\tinypm$\tiny{1.4} & \textbf{\underline{85.0}}$\tinypm$\textbf{\underline{\tiny{2.2}}} & \underline{69.8}$\tinypm$\tiny{3.3} & \underline{\textbf{73.9}}$\tinypm$\tiny{\underline{\textbf{2.3}}}  \\

		\bottomrule
	\end{tabular}
	\caption{Comparison of active learning methods. Random,
		BADGE,
		\almethod (inter-prompt uncertainty sampling
		with diversity)
		are non-deterministic. We run these algorithms for five
		random seeds and then
		average accuracy and standard deviation (averaged
across training sets with a single trial
		for non-deterministic algorithms). Best results are indicated in bold, results better than random are underlined.}
	\label{tab:al_results}
\end{table*}

Table \ref{tab:al_results} compares our AL algorithms with 
uncertainty and diversity-based prior-work.
To provide more stable results and fair comparison by
reducing noise from different runs, we employ
\emph{multiprompt finetuning with
	ENSEMBLE\textsubscript{pred}} for each
AL algorithm in this section.
Our results show that all uncertainty-only
algorithms -- entropy, lowest confidence, breaking ties and
\ppkl\ (\linefont{L2}-\linefont{L5}) -- perform worse than 
random selection (\linefont{L1}) on the
average over five datasets.
Our interpretation is that, considering that we are
finetuning a PLM with few examples, finetuning with the
highest uncertainty examples
does not generalize well.
In contrast,
\citet{schroder-etal-2022-revisiting} found that
uncertainty-only AL consistently
performs better than random selection for
fully supervised settings in PLMs.

AL prior-work that combines
uncertainty and diversity --  CAL (\linefont{L6})
and BADGE (\linefont{L7}) -- perform better than uncertainty-only
algorithms (\linefont{L2}-\linefont{L5}). Furthermore, BADGE outperforms random 
on three out of five tasks.
However,
BADGE has higher standard deviation (3.3) than random
(2.8).

Finally, when averaged over the five tasks, our proposed algorithm \almethod (\linefont{L8}) performs
better than random (\linefont{L1}) and better 
than all AL prior-work (\linefont{L2}-\linefont{L7})
with higher accuracy and lower standard
deviation.

\section{Analysis}
We now perform an in-depth analysis of AL
algorithms to understand their relative
performance better and to understand failure cases
like SST-5. We believe that these insights
will lead to improved AL strategies in future
work.

\paragraph{Balancing desiderata in AL}
\label{sec:al_additional_analysis}
We investigate three desiderata in AL:
\emph{diversity}, \emph{representativeness} and \emph{label entropy}.

Diversity \cite{zhdanov2019diverse}
measures the redundancy/similarity of training examples
by calculating the reciprocal of the average distance between
unlabeled examples and their nearest training example.
Representativeness \cite{dor2020active} captures the
well-known issue of selecting outlier examples in AL; it
is calculated as the reciprocal of the average distance
between the selected training examples and their $k$
($k$=10) nearest neighbors from unlabeled examples.  Label
Entropy \cite{prabhu-etal-2019-sampling} is the KL
divergence between the class distribution of the unlabeled
data and that of the selected training examples.

\begin{table}[t]
	\footnotesize
	\setlength\tabcolsep{4.6pt}
	\centering
	\begin{tabular}{lccccc}
		& {\textbf{Acc. $\uparrow$}} & {\textbf{Rank $\downarrow$}} & {\textbf{Div. $\uparrow$}} & {\textbf{Repr.$\uparrow$}} & {\textbf{Ent.$\downarrow$}} \\
		\midrule	
		Random & 72.6{$\tinypm$\tiny{2.8}} & 4.0 & \textbf{13.6} & \textbf{17.6} & \textbf{2.0} \\
		\midrule
		Entropy & 70.9 & 6.4 & 13.3 & 16.9 & 6.1 \\
		LC & 70.9 & 5.6 & 13.5 & 17.2 & 5.3  \\
		BT & 72.1 & 4.0 & 13.4 &  17.1 & 5.6 \\
		\PPKLshort & 69.1 & 5.6 & 13.4 & 16.9 & 9.0 \\
		CAL  & 70.4 & 4.4 & 13.1 & 17.1 & 23.5 \\
		BADGE & 73.2$\tinypm$\tiny{3.3} & \textbf{3.0} & \textbf{13.6}  & \textbf{17.6} & 2.2 \\
		\textbf{\almethod} & \textbf{73.9}$\tinypm$\tiny{\textbf{2.3}} & \textbf{3.0} &13.5 & \textbf{17.6} & \textbf{2.0} \\
		\bottomrule
	\end{tabular}
	
	\caption{Comparing average accuracy, ranking,
		diversity (D), representativeness (R), 
		and label entropy (E) scores for seven active learning algorithms.
		\almethod outperforms AL
		algorithms by better balancing out the three
		desiderata D, R and E.
Uncertainty-only
		AL algorithms have worse scores
		for each of D, R, E, resulting in lower performance.
		Arrows indicate whether higher ($\uparrow$)
		or lower ($\downarrow$) is better.}  \label{tab:al_ranking}
\end{table}

Table \ref{tab:al_ranking} shows that our proposed AL
algorithm, \almethod, outperforms all AL
prior-work with higher average accuracy, better ranking, and lower standard
deviation. It outperforms random,
a strong baseline, by 1.3 points.

Table \ref{tab:al_ranking} also shows that uncertainty-only
algorithms usually have higher label entropy (E) -- which
indicates a class distribution different from unlabeled
data -- and lower representativeness (R) and diversity (D).
We see that random selection provides a strong
baseline for E, R and D.
While BADGE and \almethod have similar scores for
E, R and D, a slight change in label entropy
(i.e., selecting a training set with different distribution than unlabeled data)
can cause a big performance drop. Our algorithm
\almethod outperforms prior-work AL because it
achieves a good balance between  the three desiderata of a good
few-shot training set:
diversity, uncertainty and representativeness.

\paragraph{\almethod: Underlying assumptions}
\label{sec:al_limitations}
We observe
that a training \dataissue strategy for few-shot finetuning
is
quite challenging
because it can only rely on zero-shot information
gathered from PLMs -- in contrast to
the fully-supervised
PLM setting. Therefore, we now share our insights on the limitations
of \almethod on SST-5 and MRPC   to guide future
work in the field.

Table \ref{tab:al_results} shows that \almethod performs
worse than random only on SST-5 even though there is no 
large difference between random and \almethod
for
diversity, representativeness, and label entropy
on
SST-5 (not shown in 
the table). The problem is that the SST-5 classes
negative/very negative and positive/very positive
are not clearly differentiated.
In a manual investigation, we found that
\almethod
selects examples that are good candidates either for both
negative/very negative or for both positive/very positive.
Thus, \almethod succeeds in identifying the most challenging
examples. But training on these does not increase accuracy
because this is an underlying uncertainty of the dataset.
To test this, we finetuned a fully-supervised
RoBERTA\textsubscript{LARGE} model on a fine-grained
sentiment analysis task with
YELP \cite{zhang2015character}. This model achieves $47.7\%$
accuracy on randomly selected examples vs.\
$39.2\%$ with \almethod. This suggests
that \almethod selects  challenging examples (i.e., not
clear which class they belong to), which are non-helpful
examples if there is underlying uncertainty in class distinctions.

In summary, \almethod makes the assumption that
discrimination between
classes can be learned well. If that is not the case, then
it can underperform.

Table \ref{tab:al_results} also illustrates a rather small
improvement in MRPC with a higher standard deviation than
random. MRPC unlabeled data  have a
non-uniform distribution
68:32 for equivalent class vs non-equivalent class.
As indicated by its
label entropy score (2.0), 
\almethod usually selects training sets with a distribution similar to
the original --  because of its clustering
mechanism.
However, \almethod selected a training set with a
53:47 distribution
in one of the five selections -- very different from 
 68:32 and resulting in low accuracy (64.2, not
 shown).
\almethod's  four other  selections are close to
 68:32 and have
higher accuracy (71.2$\pm$1.3).

In summary, \almethod makes the assumption that
selected training sets have a label distribution similar to
the overall distribution. If this assumption is not true, it
can underperform.

\begin{table}[t]
\footnotesize
\centering
	\setlength\tabcolsep{3.2pt}
	\begin{tabular}{lcc}
		& {\textbf{ALBERT}} & {\textbf{RoBERTa}}\\
		\midrule
		\ourmethod & \textbf{73.9} & \textbf{72.7} \\
		w/o active learning  & 72.6 & 71.8 \\
		w/o ensemble  & 72.0 & 71.1 \\
		w/o multiprompt & 71.6 & 70.7 \\
		\bottomrule
	\end{tabular}
	
	\caption{Ablation.
       Performance over the five tasks for
       ALBERT\textsubscript{xxlarge-v2} and
       RoBERTa\textsubscript{LARGE}. Average of five runs.
MEAL: our method with \almethod (active learning),
ENSEMBLE\textsubscript{pred} and multiprompt.
The rows show cumulative  performance drop -- so the last row
       corresponds to random selection, no ensemble, single
       prompt.
       The ablation demonstrates that each component
       contributes to MEAL's overall performance gain.
}  \label{tab:ablation_study}
\end{table}

\smallskip

Table
\ref{tab:ablation_study} shows an \textbf{ablation study}
that looks at \ourmethod's three main components: active
learning, ensemble, multiprompting.
We see that,
in addition to providing more stable results, \ourmethod increases
overall performance by 2.3 and 2.0 points over default
prompt-based finetuning for ALBERT \cite{Lan2020ALBERT}
and RoBERTa$_{\text{LARGE}}$ \cite{roberta}.
The AL module of \ourmethod, \almethod, gives the highest
performance improvement by $1.3$ and $0.9$ points (``w/o
active learning''),
showing the potential of AL in few-shot learning.
ENSEMBLE\textsubscript{pred} improves overall
performance by $0.6$ and $0.7$ (``w/o ensemble''); recall that apart from this
positive performance effect, ensembling has the added benefit 
of improved stability 
(see \S\ref{sec:res_instability}). We see an
additional improvement
 with multiprompt finetuning by $0.4$  (``w/o multiprompt''), which means
 that
 multiprompt is a win both on accuracy and 
on reduced  model space complexity.
 Finally, we see consistent
 improvements for each component across the two PLMs; this
 illustrates \ourmethod's robustness.

\section{Conclusion}
\label{sec:conclusion}
We demonstrate two stability problems of few-shot
classification with prompt-based finetuning: instability due
to \runissue and training \dataissue. We
 show that
existing solutions for instability fail.
We first propose finetuning a single 
model with multiple prompts.
This results in better performance and 
less model space complexity than
finetuning several models with single prompts.
We then propose run ensemble techniques 
that improve stability
and overall performance.

Our setup with less \runissue allows us
to explore training \dataissue for prompt-based finetuning
in a sufficiently stable experimental setup.
We compare a set of modified AL
algorithms to reduce training \dataissue
instability and improve overall performance. Our novel AL
algorithm, inter-prompt uncertainty sampling
with diversity (\almethod), outperforms prior AL algorithms
(and random selection) for both ALBERT and
RoBERTa$_\text{LARGE}$.

Apart from our algorithmic innovations for few-shot
prompt-based learning,
we hope that our study 
will support 
fairer comparison of algorithms and thereby help better
track progress in NLP. We publicly share our code and data splits
in \url{https://github.com/akoksal/MEAL}.

\section*{Acknowledgements}
This work was funded by Deutsche Forschungsgemeinschaft
(project SCHU 2246/14-1).
We would like to thank Esin Durmus, Leonie Weissweiler, and Lütfi Kerem Şenel for their valuable feedback.

\bibliography{anthology,custom}
\bibliographystyle{acl_natbib}

\appendix

\section{Appendix}
\label{sec:appendix}

\subsection{Limitations}
We exploit information in Pretrained Language Models (PLMs) to effectively use them in prompt-based finetuning for few-shot classification. Therefore prompt-based few-shot classification might be more open to reflecting biases in pretrained language models. Furthermore, our ENSEMBLE\textsubscript{pred} technique increases the model complexity during inference time to provide better performance and stability. However, ENSEMBLE\textsubscript{para} can be an alternative as it also improves performance and stability over the default setting while maintaining model complexity same. Furthermore, this work is only demonstrated for tasks in the English language.

\subsection{Implementation Details}
\label{sec:appendix_implementation}
We implement our contributions on top of the PET library. \footnote{\href{https://github.com/timoschick/pet}{https://github.com/timoschick/pet}} For prompt-based finetuning with single and multiple prompts, we use 10 epochs, 1e-5 learning rate, and also the default parameters for other hyperparameters. For a longer training approach, we use 50 epochs with a 1e-6 learning rate and warm-up with a ratio of 0.1. We report results with five different runs on five different training data. 

We conduct our experiments with NVIDIA GeForce GTX 1080 Ti, and one run of prompt-based finetuning with a single prompt takes approximately 20 minutes while one run of prompt-based finetuning with a multiprompt approach takes approximately 18 minutes for 32 examples and 4 prompts via ALBERT\textsubscript{xxlarge-v2} model with 223M parameters. As described in \S \ref{sec:active_learning}, we select training sets with $16*L$ examples where $L$ is the number of labels. Therefore, RTE, SST-2, and MRPC have 32 training examples, SST-5 has 80 training examples, and TREC has 96 training examples. RTE, SST-2, SST-5, TREC, and MRPC contain 277, 872, 1101, 500, and 408 validation examples respectively.

\subsection{Datasets}
\label{sec:appendix_datasets}
We provide four fixed prompts and verbalizers for each dataset without performing prompt search. RTE and MRPC prompts are taken from \cite{schick-schutze-2021-just} and SST-2 and SST-5 prompts are taken from \cite{gao-etal-2021-making}. We provide a set of prompts for TREC without performing a prompt search as we were not able to find a prior work on prompting with the TREC dataset.

\textbf{RTE} \cite{rte} is a textual entailment dataset that contains text pairs of premise and hypothesis with the objective of detecting entailment and contradiction. For a premise-hypothesis pair ($p$, $h$), we use

\[\pattern{$h$\textsf{\small?$\,|\,$\mask{}, }$p$}, \pattern{\textsf{\small``}$h$\textsf{\small''?$\,|\,$\mask{}, ``}$p$\textsf{\small''}},\]
\[\pattern{$h$\textsf{\small?$\,|\,$\mask{}. }$p$},
\pattern{\textsf{\small``}$h$\textsf{\small''?$\,|\,$\mask{}. ``}$p$\textsf{\small''}} \]

\noindent patterns and a verbalizer \textsf{\small yes} and \textsf{\small no} for entailment and no entailment labels.

\textbf{SST-2} and \textbf{SST-5} \cite{sst} is a sentiment
analysis dataset of movie reviews. SST-2 contains two
classes: positive and negative. SST-5 has five labels;
very positive, positive, neutral, negative, and very
negative. For a given movie review $t$, we use
\[\pattern{$t$ \textsf{\small It was \mask{}.}}, \pattern{$t$ \textsf{\small A \mask{} one.}},\]
\[\pattern{$t$ \textsf{\small The movie is \mask{}.}},
\pattern{$t$ \textsf{\small All in all \mask{}.}}\]
\noindent patterns and verbalizers ``great'', ``terrible''
(SST-2) and  ``great'', ``good'', ``okay'', ``bad'',  ``terrible'' (SST-5).

\textbf{TREC} \cite{trec} is a question classification dataset. We use six coarse classes: abbreviation, entity, description, human, location, numeric value. For a  question $t$, we use
\[ \pattern{\textsf{\small[Question Category: \mask{}] }$t$}\text{,} \pattern{\textsf{\small[Category: \mask{}] }$t$},\]
\[ \pattern{$t$ \textsf{\small This question is related to \mask{} category.}},\]
\[ \pattern{\textsf{\small I'd like to ask a question about \mask{}. } $t$}\]

\noindent patterns and verbalizers ``abbreviated'', ``entity'', ``description'', ``human'', ``location'', ``number''.

\textbf{MRPC} \cite{mrpc} is a paraphrase identification dataset of sentence pairs. For two sentences $(t_1, t_2)$, the task is to decide whether they are semantically equivalent or not. We use the same pattern and verbalizer as for RTE.

\subsection{\almethod Algorithm}
We provide the pseudo-code of \almethod in Algorithm \ref{alg:algorithm}.
\label{sec:appendix_algorithm}
\begin{algorithm*}
	\caption{IPUSD: Inter-prompt uncertainty sampling with diversity}\label{alg:algorithm}
	\begin{algorithmic}[1]
		\Require Pretrained Language Model $M$, Unlabeled Set
		$U$, Size of Training Set $N$, Pattern Set $\mathcal{P}$
		\State Compute logits representation for each example in the unlabeled set, $L(u)$ by concatenating each pattern's logits: $M(X=x_p)$ for $x\in U$ and $p\in \mathcal{P}$  
		\State Train k-means from logits representation with 8 clusters
		\For{\texttt{iter}\ $=1,2,...,1000$}
		\State Select $N$ examples, uniformly distributed in $8$ clusters with random seed \texttt{iter}, called $X_{\texttt{iter}}$
		\State \texttt{Scores\textsubscript{iter}}\ $ = \sum\limits_{x_i\in X_{\texttt{iter}}}\sum\limits_{(p,\ q)\in \mathcal{P}^{2}}\mbox{KL}(P(y|x_{i,p}, M)||P(y|x_{i,q},M))$
		\EndFor
		\State \texttt{MostUncertainIteration} $= \underset{\text{\texttt{iter}}=1,2,...,1000}{\mathrm{\textbf{argmax}}}$ \texttt{Scores\textsubscript{iter}}
		\State \textbf{return} $X$\textsubscript{\texttt{MostUncertainIteration}}
	\end{algorithmic}
\end{algorithm*}

\end{document}